\PassOptionsToPackage{table}{xcolor}
\documentclass[sigconf]{acmart}

\usepackage{booktabs} 

\usepackage{amsmath}
\usepackage{amssymb}
\usepackage{graphicx}
\usepackage{multirow}
\usepackage{subcaption}
\usepackage{enumitem}

\usepackage[font={bf}]{caption}
\usepackage[ruled,linesnumbered,boxed]{algorithm2e}
\usepackage{color, url}
\usepackage{xspace} 
\usepackage{balance}

\usepackage{hyperref}
\hypersetup{%
  colorlinks=true,%
  linkcolor= blue,
  citecolor= blue,
  urlcolor= blue,
  linkbordercolor=white,
  urlbordercolor= blue,
  citebordercolor=white,
  pdfborder={0 0 1},%
}

\newcommand{\eat}[1]{}
\newcommand{\ie}{\emph{i.e.}, }
\newcommand{\etal}{\emph{et al.} }
\newcommand{\eg}{\emph{e.g.}, }

\usepackage{threeparttable}

\usepackage{tabu}

\usepackage{ctable}
\usepackage{array}
\usepackage{arydshln}
\usepackage[table]{xcolor}
\newcommand{\thline}{\specialrule{.1em}{.1em}{.1em}}
\newcolumntype{L}[1]{>{\raggedright\let\newline\\\arraybackslash\hspace{0pt}}m{#1}}
\newcolumntype{C}[1]{>{\centering\let\newline\\\arraybackslash\hspace{0pt}}m{#1}}
\newcolumntype{R}[1]{>{\raggedleft\let\newline\\\arraybackslash\hspace{0pt}}m{#1}}

\newenvironment{packeditemize}{\begin{list}{$\bullet$}{\setlength{\itemsep}{0.2pt}\addtolength{\labelwidth}{-4pt}\setlength{\leftmargin}{\labelwidth}\setlength{\listparindent}{\parindent}\setlength{\parsep}{1pt}\setlength{\topsep}{2pt}}}{\end{list}}

\setcopyright{rightsretained}

\acmDOI{10.475/123_4}

\acmISBN{123-4567-24-567/08/06}

\acmConference[Conf'18]{ACM  conference}{July 2018}{City, State USA} 
\acmYear{2018}
\copyrightyear{2018}

\acmPrice{15.00}

\begin{document}

\setlength{\paperheight}{11in}
\setlength{\paperwidth}{8.5in}
\setlength{\pdfpageheight}{\paperheight}
\setlength{\pdfpagewidth}{\paperwidth}
\setlength{\abovedisplayskip}{3pt}
\setlength{\belowdisplayskip}{3pt}

\title{Simultaneous Modeling of Multiple  Complications for Risk Profiling in Diabetes Care}

\author{Bin Liu$^{1}$, Ying Li$^{1, 2}$, Soumya Ghosh$^{1, 2}$, Zhaonan Sun$^{1}$,  Kenney Ng$^{1, 2}$, Jianying Hu$^{1, 2}$}
\affiliation{%
  \institution{1: Center for Computational Health, IBM Research \hskip 1cm
  2: MIT-IBM Watson AI Lab}
}
\email{bin.liu1@ibm.com, {liying, ghoshso, zsun, kenney.ng, jyhu}@us.ibm.com}

\begin{abstract}
Type 2 diabetes mellitus (T2DM) is a chronic disease that often results in multiple complications.
Risk prediction and profiling of T2DM complications is critical for healthcare professionals to design personalized treatment plans for patients in diabetes care for improved outcomes. 
In this paper, we study the risk of developing complications after the initial T2DM diagnosis from  longitudinal patient records.
We propose a novel multi-task learning approach to simultaneously model multiple complications where each task corresponds to the risk modeling of one complication. 
Specifically, the proposed method strategically captures the relationships
(1) between the risks of multiple T2DM complications,
(2) between the different risk factors, and 
(3) between the risk factor selection patterns.
The method uses coefficient shrinkage to identify an informative subset of risk factors from high-dimensional data, and uses a hierarchical Bayesian framework to allow domain knowledge to be incorporated as priors.
The proposed method is favorable for healthcare applications because in additional to improved prediction performance, relationships among the different risks and risk factors are also identified.
Extensive experimental results on a large electronic medical claims database show that the proposed method outperforms state-of-the-art models by a significant margin. 
Furthermore, we show that the risk associations learned and the risk factors identified lead to meaningful clinical insights.
\vspace{-3pt}
\end{abstract}
\vspace{-15pt}

%
%

\vspace{-4pt}

\keywords{\vspace{-1pt} Diabetes Care; Complications; Risk Profiling; Multi-task Learning\vspace{-0.5pt}}

\renewcommand{\shortauthors}{B. Liu et al.}

\maketitle

\def \Pr {{\mathrm{Pr}}}

\def \xx {{\bf x}}
\def \XX {{\bf X}}

\def \ww {{\bf w}}
\def \WW {{\bf W}}

\def \uu {{\bf u}}
\def \UU {{\bf U}}

\def \uww { \underline{\bf w}}
\def \oww { \overline{\bf w}}

\def \YY {{\bf Y}}

\def \RR {{\bf R}}

\def \II {{\bf I}}

\vspace{-5pt}
\section{Introduction}
Type 2 diabetes mellitus (T2DM) is a chronic disease that affects nearly half a billion people around the globe~\cite{world2016global}. 
T2DM is characterized by hyperglycemia--- abnormally elevated blood glucose (blood sugar) levels, and is almost always associated with a number of complications~\cite{forbes2013mechanisms}. Over time, the chronic elevation of blood glucose levels caused by T2DM leads to blood vessel damage which in turn leads to associated complications, including kidney failure, blindness, stroke, heart attack, and in severe cases even death. 
Meanwhile, the cost of diabetes care has been increasing over the past decades and the annual cost is a staggering~\cite{CDC:diabetes:2017,ncd2016worldwide}.
T2DM management requires continuous medical care with multifactorial risk-reduction strategies beyond glycemic control~\cite{american2013standards}. Risk profiling of T2DM complications is critical for healthcare professionals to appropriately adapt personalized treatment plans for patients in diabetes care, improving care quality and reducing cost. 

The recent abundance of the electronic health records (EHRs) and electronic medical claims data has provided an unprecedented opportunity to apply predictive analytics to improve T2DM management. In this paper, we study the risk profiling of T2DM complications from longitudinal patient records: {\it what is the probability that a patient will develop complications within a time window after the initial T2DM diagnosis?}  In the literature, EHRs and claims data have been leveraged for a wide range of healthcare applications~\cite{Kenney2016early,razavian2015population,himes2009prediction,cheng2016risk,choi2017using,wang:progression:kdd2014,wang2015towards,chen2016patient,he2014mining,tabak2013using,Nori:KDD2015}. However, there are unique difficulties that arise when performing data-driven risk prediction and profiling of T2DM complications from patient medical records: 
\begin{packeditemize}
\vspace{4pt}
\item The first challenge stems from the need to effectively capture correlations between multiple T2DM complications. Considering that the different complications are manifestations of a common underlying condition--hyperglycemia, modeling complications as independent of one another leads to suboptimal models.
\item Patient medical record data contain rich information about relationships among medical concepts and risk factors, pertinent to T2DM. However, developing statistical methods that can effectively exploit this information is challenging.   
\item Further, when using patient medical record data for risk prediction and profiling, each patient is typically represented by very high-dimensional data while only a small subset of the predictors are actually relevant. It is essential to be able to identify the subset of predictors that are useful for predictive analysis to facilitate model transparency and interpretability.
\item Finally, it is desirable for the model to have the ability to leverage T2DM domain knowledge. Such clinical domain knowledge is often available or partially available, and incorporating it into the analysis can lead to more accurate inferences.
\vspace{4pt}
\end{packeditemize}

In this paper, we address these challenges by developing methods for simultaneously modeling multiple complications for risk profiling in diabetes care.  We begin by formulating T2DM complication risk prediction as a Multi-Task learning (MTL)~\cite{Caruana:MLT1997} problem with each complication corresponding to one task. MTL jointly learns multiple tasks using a shared representation so that knowledge obtained from one task can help the other tasks.
We then develop extensions that in addition to capturing task relationships driven by the underlying disease also model dependencies between information-rich features (risk factors). Further, assuming that similar T2DM complications have similar contributed risk factors, we endow our models with the ability to perform correlated shrinkage through a novel correlated Horseshoe distribution. This allows us to identify subsets of risk factors for different complications while accounting for associations between complications. %
We call the proposed method {\bf T}ask {\bf RE}lationship and {\bf F}eature relationship {\bf Le}arning with correlated {\bf S}hrinkage (TREFLES).
We formulate TREFLES in a hierarchical Bayesian framework, allowing us to easily capture domain knowledge through carefully chosen priors.

Finally, we assess our proposed innovations through extensive experiments on patient level data extracted from a large electronic medical claims database. The results show that the proposed approach consistently outperforms previous models by a significant margin and demonstrate the effectiveness of the simultaneous modeling framework over modeling each complication independently.
Furthermore, we show that the risk associations learned and the risk factors identified lead to meaningful clinical insights.

In summary, our key contributions are as follows:
\begin{packeditemize}
\item We provide a systematic study on risk profiling by simultaneously modeling of multiple complications in chronic disease care using T2DM as a case study. 
\item We design a novel model, \emph {TREFLES}, that jointly captures relationships between risks, risk factors, and risk factor selection learned from the data with the ability to incorporate domain knowledge as priors.
\item We demonstrate the effectiveness of TREFLES in both predictive capability and clinical interpretability via a comprehensive study of T2DM complications using a large electronic medical claims database.
\end{packeditemize}
The proposed method is favorable for healthcare applications beyond diabetes care. It provides a powerful tool for not only improving predictive performance, but also for recovering clinically meaningful insights about relationships among different risks and risk factors.

\section{Simultaneous Modeling of Multiple Complications for Risk Profiling}
In this section, we first formulate the problem of diabetes complications risk profiling, and then introduce the proposed approach to simultaneously model multiple complications, addressing the aforementioned challenges.

\subsection{Diabetes Complications Risk Profiling}
\begin{figure*}
\centering
        \begin{subfigure}[t]{0.67\textwidth}
                \centering\includegraphics[width=1.05\textwidth]{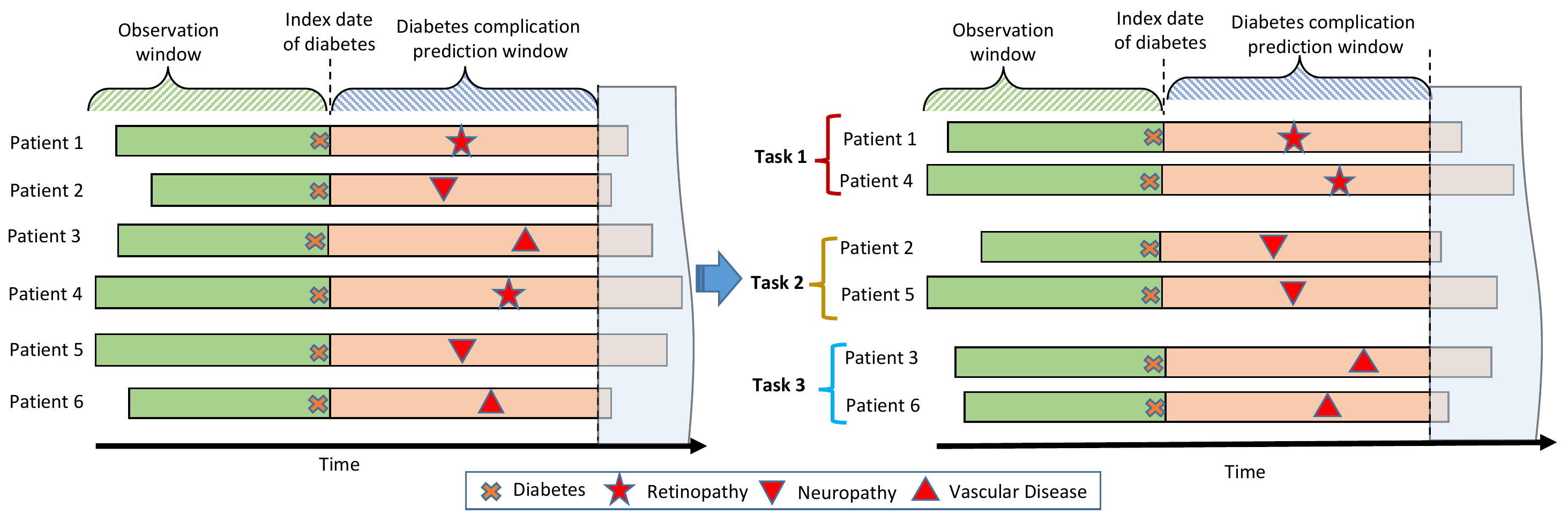}
                \caption{Multi-task learning formulation.}
                \label{fig:MTL_formulation}
        \end{subfigure}%
        \hfill
        \begin{subfigure}[t]{0.33\textwidth}
                \centering\includegraphics[width=0.8\textwidth]{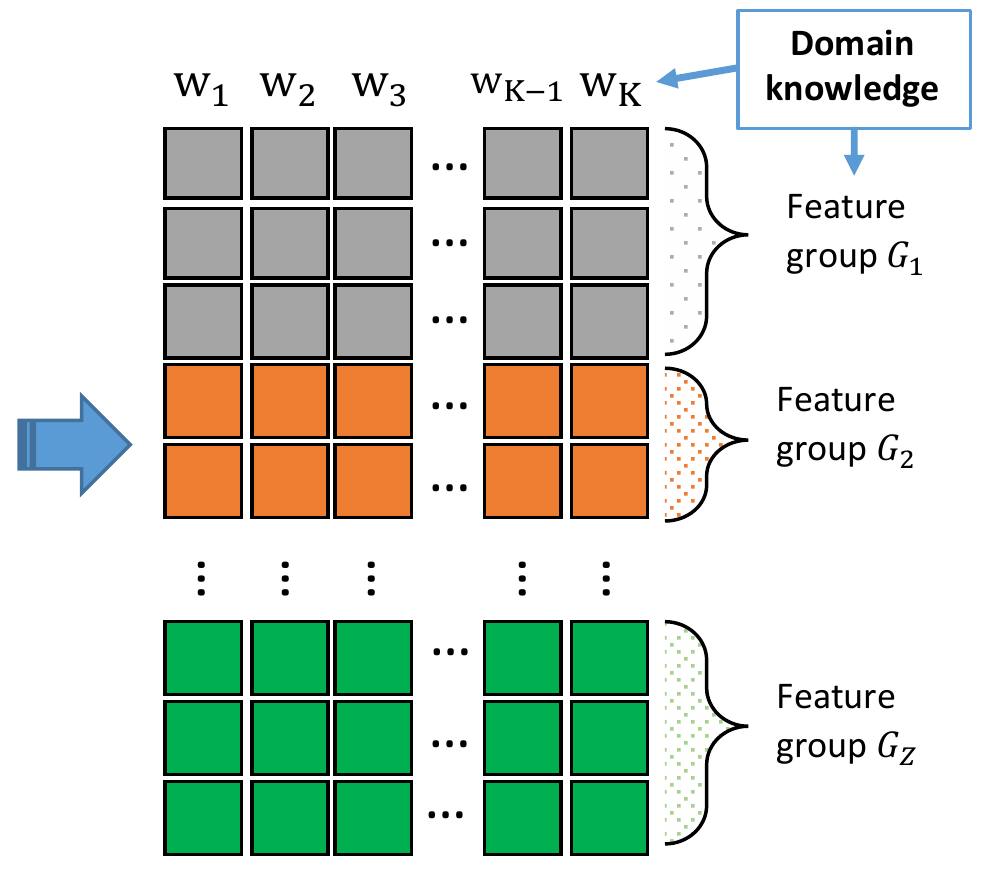}
                \caption{Coefficient matrix.}
                \label{fig:coef_matrix}
        \end{subfigure}%
\vspace{-0.3cm}
\caption{Proposed framework for simultaneous modeling of multiple T2DM complications. (a) Multi-task learning formulation: the predictions of multiple complications in consideration (\eg retinopathy, neuropathy and vascular disease) are grouped into different tasks where each task models only one complication. Multi-task learning (MTL) is applied to capture the association between the different complications. Features are derived from patients' medical records up to the time of the initial T2DM diagnosis. Outcome is evaluated in the follow-up window. To simplify the illustration, only  positive cases are shown. (b) The correlations between complication risks are revealed in the structure of the coefficient matrix $\WW$, which captures both the relationships between T2DM complication risk profiling tasks and the correlation between the features.}\label{fig:prediction_framework}
\vspace{-0.3cm}
\end{figure*}

The goal is to build an effective approach to predict the risk of a patient developing complication(s) within a follow-up window $\Delta t$ after the initial T2DM diagnosis. Specifically, as shown in Figure \ref{fig:prediction_framework}, for each patient $i\in\{1, \dots, N\}$ we observe a set of $M$ features (risk factors), denoted as $\xx_i=[x_{i1}, x_{i2}, \dots, x_{iM}]^\top$, for an observation window up until the patient was initially diagnosed with T2DM. Let there be $K$ complications in consideration indexed by $k\in\{1, \dots, K\}$. We use $c_{ki}\in \{0,1\}$ to represent the onset event of patient $i$ developing complication $k$ in the follow-up window $\Delta t$ and use $y_{ki}$ to represent the event probability (risk).  For each task $k$ we observe a set of complications $\mathcal D_k=\{x_i, c_{ki}\}_{i\in \mathcal N_k}$, where $\mathcal N_k$ are the patients observed in complication $k$. The set of all observed complication events are denoted as $\mathcal D= \{\mathcal D_k\}_{k=1}^K$. Given $\mathcal D$, we aim to build a predictive model $y_{ki}=\Pr(c_{ki}|\Theta, \xx_i)$, where $\Theta$ are the model parameters, to predict the risk that patient $i$ will develop complication $k$ during follow-up. Table \ref{table:math_notation} summarizes useful notation used in the remainder of the paper. 
\begin{table}
\caption{Mathematical Notations}\label{table:math_notation}
\vspace{-0.5cm}
\addtolength{\tabcolsep}{-4pt}
\begin{center}
\rowcolors{2}{gray!25}{white}
    \begin{tabular}{  L{1.8cm}  L{6.4cm}}
    \rowcolor{gray!50}
    \thline\hline
    Symbol  & Description \\ \hline
    $N, M, K$ & number of subjects,  features, and complications\\
    $i, j, k$ & index of subjects, features, and complications\\
    $c_{ki}\in \{0,1\}$ & event of complication $k$ for patient $i$ where $1$ indicates an observed event and $0$ otherwise  \\
    $y_{ki}$ & probability (risk) of  patient $i$ for complication $k$ \\
    $\xx_i \in \mathbb{R}^M$ & vector of features for patient $i$ \\
    $\ww_k \in \mathbb{R}^M$ &  vector of coefficients for complication $k$\\
    $\WW \in \mathbb{R}^{M\times K}$ & $\WW = [\ww_1, \cdots, \ww_K]$ is the matrix of coefficients \\ 
    $\ww^j \in \mathbb{R}^K$ &  $\ww^j$ is the $j^{\mathrm {th}}$ row of  coefficient complication $\WW$\\
    $\mathbf{\Omega} \in \mathbb{R}^{K\times K}$ &  matrix of relatedness  between  complications\\
    $\mathbf{\Omega}_0 \in \mathbb{R}^{K\times K}$ & matrix of prior knowledge about risk association\\
    $z, \mathcal{G}_z$ & index and the $z^{\mathrm {th}}$ group of features\\
    $\mathbf{\Sigma}_z \in \mathbb{R}^{G_z\times G_z}$ & correlation matrix between features in group $G_z$\\
    \hline\thline
    \end{tabular}
\end{center}
\vspace{-0.5cm}
\end{table}

\vspace{-5pt}
\subsection{Learning Associations between Multiple Complications}
Given the features (risk factors) $\xx_i=[x_{i1}, x_{i2}, \dots, x_{iM}]^\top$ observed up until the initial T2DM diagnosis for patient $i$, we model the risk of patient $i$ developing complication $k$ in the follow-up window $\Delta t$ as:
\begin{equation}\label{stl_model}
y_{ki} = \Pr(c_{ki}|\Theta, \xx_i) = \sigma(\ww_k^\top \xx_i)
\end{equation}
where $\ww_k$ is the coefficient vector for complication $k$, and $\sigma (t)$ is a logistic function $\sigma (t)={\frac {1}{1+e^{-t}}}$. Then the event onset can be modeled as a draw from a Bernoulli distribution $c_{ki} \sim \mathrm{Bernoulli}(\sigma(\ww_k^\top \xx_i))$.

To capture and leverage the association between the risks of the different T2DM complications, we formulate the complication risk prediction problem as a multi-task learning problem. As shown in Figure~\ref{fig:prediction_framework}, we group the predictions of multiple complications in consideration (e.g., retinopathy, neuropathy and vascular disease) into different learning tasks. Each task models only one complication risk via Equation (\ref{stl_model}). 
Next, we apply multi-task learning to capture the association between different complications. 

\subsection{Learning Multi-task and Feature Relationships with Correlated Shrinkage} 

We aim to capture three types of dependencies in our framework. First, the complication tasks are related since they all stem from a common underlying condition--hyperglycemia. Second, there are associations between the features since they are derived from and represent the health status of the same set of real patients. Third, similar T2DM complications have similar contributing risk factors that lead to the development of those complications.

\subsubsection{Modeling Task and Feature Associations}
~\\Let $\WW = [\ww_1, \ww_2, \dots, \ww_K] \in \mathbb{R}^{M\times K}$ denote the matrix of coefficients of all $K$ complications in consideration. To explore the latent association between the risks of T2DM complications,  we impose explicit structure on the coefficient matrix $\WW$.  
Specifically, we assume that the coefficient matrix $\WW$ follows a Matrix Variate Normal (MVN) distribution:
\begin{equation}\label{Equ:MVN}
\boldsymbol{\WW} \sim \mathcal{MVN} (\mathbf{0}, \mathbf{\Sigma}, \mathbf{\Omega}).
\end{equation}
The the first term $\boldsymbol{0}$ is a $M \times K$ matrix of zeros representing the mean of $\WW$. The second term $\mathbf{\Sigma}$ is a $M \times M$ symmetric positive definite matrix representing the row-wise covariances of $\WW$, {\it i.e.} the correlations between the features. 
The third term $\mathbf{\Omega}$ is a $K \times K$ symmetric positive definite matrix representing the column-wise covariance of $\WW$, {\it i.e.} the correlations between the tasks.

Equation (\ref{Equ:MVN}) captures both the relationships between tasks through $\mathbf{\Omega}$ and correlations among features through $\mathbf{\Sigma}$. As a result, this formulation is a generalization~\cite{zhao2017FETR} of the two most widely used MTL strategies: the task relation learning approaches~\cite{Zhang:MLT:UAI2010,Zhang:MTL:TKDD2014} and the feature relationship learning approaches~\cite{Argyriou:MLFT:nips2007,Argyriou:MLFT:MLJ2008}. 
When $\mathbf{\Sigma}$ is diagonal, we recover task relationship learning, and 
by setting $\mathbf{\Omega}$ to a diagonal matrix, we recover feature relationship learning.

In healthcare, features can be very fine-grained and domain knowledge is often available to group similar features into higher level representations.
In this paper, we leverage this knowledge and group the diagnosis features in the patient medical records according to the ontologies of the International Classification of Diseases (ICD)~\cite{world1978:ICD9}.  
As a result, we group the features $\{x_j\}_{j=1}^M$ into $Z$ groups $\{\mathcal G_z\}_{z=1}^Z$, where $\mathcal G_z$ has $G_z$ features with $\sum_z G_z = M$.
Let $\ww^j = [w_{j1}, w_{j2}, \dots, w_{jK}] \in \mathbb{R}^K$ be the $j$ row of coefficient complication matrix $\WW$, then $\ww^j$ is the $j^{\mathrm {th}}$ coefficient across the $K$ tasks. As shown in Figure \ref{fig:coef_matrix}, we group coefficient matrix $\WW$ into $Z$ blocks where each $\WW_z \in \mathbb{R}^{G_z \times K}$ is a matrix block where feature $j$ belongs to group $\mathcal G_z$, namely, $\WW_z= \{\ww^j\}_{j\in \mathcal G_z}$. We assume $\WW_z$ follows a Matrix Variate Normal (MVN) distribution:
\begin{equation}\label{Equ:structure}
\begin{split}
\WW_z & \sim \mathcal{MVN} (\mathbf 0, \mathbf{\Sigma}_z, \mathbf{\Omega})\\
\end{split}
\end{equation}
where ${\mathbf 0}$ is the mean, $\mathbf{\Sigma}_z$ is the correlations between features,  and $\mathbf{\Omega}$ is the correlations between tasks. The zero mean indicates \emph{a-priori} the features are assumed to have no effect.  As a result, Equation (\ref{Equ:structure}) captures both the relationships between T2DM complications and the relationships between features.
Then we have,
\begin{equation}
\Pr(\WW_z | \mathbf{0}, \mathbf{\Sigma}_z, \mathbf{\Omega}) = \frac{\exp\left( -\frac{1}{2} \mathrm{tr}\left[ \mathbf{\Omega}^{-1} \WW_z^\top \mathbf{\Sigma_z}^{-1} \WW_z  \right] \right)}{(2\pi)^{KG_z/2}  |\mathbf{\Sigma_z}|^{K/2} |\mathbf{\Omega}|^{G_z/2} }.
\end{equation}

\vspace{-5pt}
\subsubsection{Correlated Shrinkage} 
EHR data is usually high-dimensional with a large numbers of potentially relevant features. 
We are interested in identifying an informative subset of coefficients, which reflect the contributing risk factors responsible for the development of a specific complication, by shrinking irrelevant coefficients towards zero. 
Sparsity-promoting priors are widely used in this context. Perhaps, the most popular example is the Laplacian prior which gives rise to the Lasso~\cite{Tibshirani:Lasso} $\ell_1$ regularizer. However, such a prior provides uniform shrinkage --- it shrinks values close and far from zero alike. The Horseshoe prior \cite{carvalho2009horseshoe,carvalho2010horseshoe} provides an attractive alternative. It maintains an infinitely tall spike at zero, while exhibiting Cauchy-like heavy tails. As a consequence, it shrinks small values to zero more strongly than the Laplace prior, while its heavy tails allow some coefficients to escape completely un-shrunk. This property allows the Horseshoe prior to be more robust to large signals while providing strong shrinkage towards zero to noise.
We can place a Horseshoe prior on $w_{jk}$ to promote sparsity on the $j^{\mathrm {th}}$ coefficient of task $k$ by setting,
\begin{equation}\label{Equ:HS}
\begin{split}
w_{jk}|\lambda_{jk}, \tau  \sim \mathcal{N} (0, \lambda_{jk}^2 \tau^2), 
\lambda_{jk}  \sim  \mathrm C^+ (0, 1), \tau  \sim  \mathrm C^+ (0, 1)
\end{split}
\end{equation}
where $\mathrm C^+ (0, 1)$ is a half-Cauchy distribution, $\lambda_{jk}$ is called the local shrinkage parameter, and $\tau$ is the global shrinkage parameter.

However, the vanilla Horseshoe prior fails to capture correlations among tasks. 
Recall that in our MTL setting, we assume that similar T2DM complications (tasks) should have similar contributing features.
Note that $\ww^j = [w_{j1}, w_{j2}, \dots, w_{jk} \dots, w_{jK}] \in \mathbb{R}^K$ is the $j^{\mathrm {th}}$ coefficient across the $K$ tasks.
Ideally, pairs of $w_{jk}, k\in\{1, \dots, K\}$ would have more similar shrinkage if their tasks ($k$) are positively correlated.

To do so, we introduce a novel \emph {correlated Horseshoe} prior. We construct the correlated Horseshoe prior by employing a Gaussian copula~\cite{Song:copulas:2009} to couple the local shrinkage parameters $\lambda_{jk}$ together via the task correlations reflected in $\mathbf{\Omega}$,  while forcing the marginals of $\lambda_{jk}$ to retain their half-Cauchy distributions.

Let $\uu^j=[u_{j1}, u_{j2}, \dots, u_{jK}] \in \mathbb{R}^K$ be a $K$-dimensional vector that follows a multivariate normal distribution 
\begin{equation}
[u_{j1}, u_{j2}, \dots, u_{jk} \dots, u_{jK}] \sim \mathcal{MN}(\mathbf 0, \mathbf{\Omega}),
\end{equation}
Observe that $ \uu^j$ preserves the correlations between tasks through $\mathbf{\Omega}$ and $u_{jk} \sim \mathcal{N}(0, \Omega_{kk})$.
Next, we need to ensure that $\lambda_{jk}$ follows the half-Cauchy distribution. We use inverse transform sampling~\cite{devroye1986sample} to guarantee half-Cauchy marginals. Inverse transform sampling is based on the result that given a uniform random variable $a \sim U(0,1)$, we can generate another random variable $b$ with a cumulative distribution function (cdf) $\mathrm F$, by setting $b = \mathrm F^{-1}(a)$, as long as $\mathrm F$ is invertible. Now, if $b\sim \mathrm C^+(0, 1)$, then $\mathrm F(b) = \frac{2}{\pi}\mathrm{tan}^{-1}(b)$ and since, $\Phi(u_{jk}) \sim U(0, 1)$, where $\Phi(u_{jk})$ is the cdf of $u_{jk}$, $\mathrm F^{-1}(\Phi(u_{jk}))$ follows a half-Cauchy distribution.
%
The correlated Horseshoe is thus completely specified as,
\begin{equation}\label{Equ:correlated_HS}
\begin{split}
&\uu^j \sim \mathcal{MN}(\mathbf 0, \mathbf{\Omega}), \quad \Phi(u_{jk}) = \frac {1}{2}\left[1+\operatorname {erf} \left({\frac {u_{jk}}{ \sqrt {2\Omega_{kk} }}}\right)\right], \\
&\lambda_{jk} = \mathrm F^{-1}(\Phi(u_{jk})) = \mathrm{tan}\left(\frac{\pi \Phi(u_{jk})}{2}\right) \quad \forall k\in\{1, \dots, K\},
 \\
&w_{jk}|\lambda_{jk}, \tau  \sim \mathcal{N} (0, \lambda_{jk}^2 \tau^2), \quad
\tau  \sim  \mathrm C^+ (0, 1).
\end{split}
\end{equation}
We emphasize that $\lambda_{jk}$s are correlated via the latent variables $\uu^j$, allowing us to preserve task correlations. At the same time their marginal half-Cauchy behavior retains the desirable properties of the Horseshoe distribution.

\subsubsection{Capturing Domain Knowledge}
In order to utilize available domain knowledge, we impose an Inverse-Wishart prior distribution 
on $\mathbf{\Omega}$
\begin{equation}
\mathbf{\Omega} \sim \mathcal{IW}( \delta \mathbf{\Omega}_0,\nu).
\end{equation}
The Inverse-Wishart distribution is a conjugate prior for the multivariate Gaussian distribution. 
$\mathbf{\Omega}_0$ is a known symmetric positive definite matrix that contains all prior knowledge about the risk associations.
$\delta$ and $\nu$ are two tuning parameters. When domain knowledge on risk associations is available, the prior distribution can leverage the information and help improve the estimation of $\mathbf{\Omega}$. When domain knowledge about risk associations is not available, we can set $\mathbf{\Omega}_0$ to be the identify matrix $\mathbf{I}$.

\subsection{Prediction}
Note that in Equation (\ref{Equ:correlated_HS}), we have $w_{jk}|\lambda_{jk}, \tau  \sim \mathcal{N} (0, \lambda_{jk}^2 \tau^2)$ and $ \lambda_{jk}$ is a function of $u_{jk}$, which is sampled from $\mathcal{MN}(\mathbf 0, \mathbf{\Omega})$. An equivalent non-centered re-parameterization is given by $\tau \lambda_{jk}\cdot w_{jk}$, where $w_{jk} \sim \mathcal{N} (0, 1)$. Here, we use this equivalent parameterization for computational convenience. Let $\mathbf{\Lambda} \in \mathbb{R}^{M\times K}$ be a matrix with element $\lambda_{jk}$, then we can reparameterize the matrix of coefficients as
$
\boldsymbol{\beta} = \tau \mathbf{\Lambda} \circ \WW,
$
where $\circ$ represents a pointwise (Hadamard) product between $\mathbf{\Lambda}$ and  $\WW$.  Finally, we model the risk of complication $k$ for patient $i$ as, $y_{ki}\mid \beta_k, \xx_i = \sigma(\boldsymbol{\beta}_k^\top \xx_i)$.

\vspace{-0.3cm}
\section{Parameter Estimation}

Let $\Theta = \left\{\{\WW_z, \mathbf{\Sigma}_z\}_{z=1}^Z, \mathbf{\Omega}, \UU, \tau\right\}$ denote all parameters to be estimated, and $\Phi=\{\mathbf\Omega_0, \delta, \nu\}$ denote all hyperparameters. For each task $k$ we observe a set of complication events $\mathcal D_k=\{\langle x_i, c_{ki}\rangle \}_{i\in \mathcal N_k}$, where $\mathcal N_k$ represents the patients observed for complication $k$. The observed complication events are denoted as $\mathcal D= \{\mathcal D_k\}_{k=1}^K$. Given $\{\mathcal D, \Phi\}$ the posterior distribution, 
\begin{equation*}
\begin{split}
& \Pr(\Theta| \mathcal D, \Phi)\\
& \propto  \Pr(\tau) \Pr(\mathbf{\Omega}) \prod_{k=1}^K\prod_{i=1}^{\mathcal N_k}\Pr(c_{ki}|\boldsymbol{\beta}_k, \xx_i) 
\prod_{z=1}^Z \Pr(\WW_z) \prod_{j=1}^M \Pr(\uu_j) \\
& \propto 
\frac{2}{\pi (1+\tau^2)} |\mathbf\Omega|^{-\frac{\nu + K + 1}{2}} \exp\left(-\frac{\delta}{2} \operatorname{tr} (  \mathbf\Omega_0 \mathbf\Omega^{-1}) \right) \prod_{k=1}^K\prod_{i=1}^{\mathcal N_k}\Pr(c_{ki}|\boldsymbol{\beta}_k, \xx_i) \\
& \times \prod_{z=1}^Z \frac{\exp\left( -\frac{1}{2} \mathrm{tr}\left[ \mathbf{\Omega}^{-1} \WW_z^\top \mathbf{\Sigma}_z^{-1} \WW_z  \right] \right)}{(2\pi )^{KG_z/2} |\mathbf{\Sigma}_z|^{K/2} |\mathbf{\Omega}|^{G_z/2} } 
 \prod_{j=1}^M \frac {\exp \left(-\frac{1}{2} \uu^j \mathbf{\Omega}^{-1} (\uu^j)^\top \right)}{ (2\pi )^{K/2}|\mathbf{\Omega}|^{1/2}}
 . 
\end{split}
\end{equation*}
We estimate the parameters via maximizing the log posterior $\ell(\Theta )= \log \Pr(\Theta| \mathcal D, \Phi)$.

\paragraph{Objective Function.} 
We rewrite the negative log-posterior $\ell(\Theta )$ to obtain the following objective function $\mathcal O  (\Theta)$ to minimize:
\begin{equation*}\label{Equ:obj_fun}
\begin{split}
\mathcal O  & =
 \sum_{k=1}^K\sum_{i=1}^{\mathcal N_k} \bigg\{ -c_{ki} \log \sigma(\boldsymbol{\beta}_k^\top \xx_i) - (1-c_{ki}) \log(1- \sigma(\boldsymbol{\beta}_k^\top \xx_i)) \bigg\}\\
& + \frac{1}{2}\sum_{z=1}^Z \bigg\{ \mathrm{tr}\left[ \mathbf{\Omega}^{-1} \WW_z^\top \mathbf{\Sigma}_z^{-1} \WW_z  \right] \bigg\}  
+  \frac{2M + K + \nu + 1}{2} \log |\mathbf\Omega|  \\
& + \frac{\delta}{2} \operatorname{tr} ( \mathbf\Omega_0 \mathbf\Omega^{-1}) 
 + \frac{K}{2} \sum_{z=1}^Z \log |\mathbf{\Sigma}_z|   
 + \frac{1}{2}\sum_{j=1}^M  \uu^j \mathbf{\Omega}^{-1} (\uu_j)^\top + 2\log(1+\tau^2) \\
\text{s.t.} & ~~\mathbf\Omega  \succeq 0, \mathbf{\Sigma}_z  \succeq 0.
\end{split}
\end{equation*}
where $\mathbf X  \succeq 0$  means that the matrix $\mathbf X$ is positive semidefinite.


Solving the above optimization problem is non-trivial. The optimization problem is not convex since $\log |\mathbf\Omega|$ and $\log |\mathbf{\Sigma}_z|$ are concave functions. Therefore we adopt an iterative algorithm to solve the problem. Within each iteration, the blocks $\WW_z$,  $\mathbf{\Sigma}_z$, $\mathbf\Omega$, $\UU$, and $\tau$ are updated alternatively. 

\vspace{3pt}
{\noindent \bf Update $\WW_z$ given others}:
With other parameters fixed, objective function w.r.t  $\WW_z$ becomes 
\begin{equation*}\label{Equ:obj_fun_W}
\begin{split}
\underset{\{\WW_z\}_{z=1}^Z}{\operatorname{arg\,min}} ~~& \sum_{k=1}^K\sum_{i=1}^{\mathcal N_k} \bigg\{ -c_{ki} \log \sigma(\boldsymbol{\beta}_k^\top \xx_i) - (1-c_{ki}) \log(1- \sigma(\boldsymbol{\beta}_k^\top \xx_i)) \bigg\} \\
 & + \sum_{z=1}^Z \Bigg\{ \frac{1}{2}\mathrm{tr}\left[ \mathbf{\Omega}^{-1} \WW_z^\top \mathbf{\Sigma}_z^{-1} \WW_z  \right] \Bigg\}.
\end{split}
\end{equation*} 
This is a convex optimization problem with respect to $\WW_z$. We use stochastic gradient descent method to update the $\{\WW_z\}_{z=1}^Z$. 
The main process involves randomly scanning training instances and iteratively updating parameters. In each iteration, we randomly sample an instance $\langle x_i, c_{ki}\rangle$, and we minimize $\mathcal O  (\Theta)$ using the update rule for 
$ \Theta = \Theta - \epsilon \cdot \frac{\partial \mathcal O  (\Theta)}{\partial \Theta}$, where $\epsilon $ is a learning rate. Note that $\ww_k = [\ww_{\mathcal G_1}, \ww_{\mathcal G_2}, \dots, \ww_{\mathcal G_Z}]^\top$ and $\WW_z= \{\ww^j\}_{j\in \mathcal G_z}$. Let $\ww_k^{\mathcal G_z} =  [w_{jk}, w_{jk}, \cdots, w_{jk}]^\top, j \in \mathcal G_z$ be the $k$ column of $\WW_z$, then $\ww_k^{\mathcal G_z}$ corresponds to the coefficients of features in group ${\mathcal G_z}$ in task $k$. Given an instance $\langle x_i, c_{ki}\rangle$, the gradient with respect to $\ww_k^{\mathcal G_z}$ is
\begin{equation}\label{update_Wz}
\begin{split}
     \frac{\partial \mathcal O }{\partial \ww_k^{\mathcal G_z}} 
     = & -\Big(c_{ki} - \sigma(\boldsymbol{\beta}_k^\top \xx_i) \Big) \frac{ \partial \boldsymbol{\beta}_k^\top \xx_i}{\partial \xx_i^{\mathcal G_z}} 
     + \left[ \mathbf{\Sigma}_z^{-1}  \WW_z \mathbf{\Omega}^{-1} \right]_{k}^{\mathcal G_z}
\end{split}
\end{equation}
where $\xx_i^{\mathcal G_z}$ is the features in group $z$, and $\left[ \XX \right]_{k}$ means the $k$-th column of matrix $\XX$. 
So we have $\frac{ \partial \boldsymbol{\beta}_k^\top \xx_i}{\partial \xx_i^{\mathcal G_z}}  = \tau \boldsymbol{\lambda}_k^{\mathcal G_z} \circ \xx_i^{\mathcal G_z}$.

\vspace{2pt}
{\noindent \bf Update $\UU$ given others}: 
With other parameters fixed, the objective function w.r.t  $\UU$ becomes 
\begin{equation*}\label{Equ:obj_fun_U}
\begin{split}
\underset{\UU}{\operatorname{arg\,min}} & \sum_{k=1}^K\sum_{i=1}^{\mathcal N_k} \bigg\{ -c_{ki} \log \sigma(\boldsymbol{\beta}_k^\top \xx_i) - (1-c_{ki}) \log(1- \sigma(\boldsymbol{\beta}_k^\top \xx_i)) \bigg\}\\
& + \frac{1}{2}\sum_{j=1}^M  \uu^j \mathbf{\Omega}^{-1} (\uu^j)^\top
\end{split}
\end{equation*}
To apply SGD, we optimize columns $\uu_k$ instead of rows $\uu^j$. Note than $\sum_{j=1}^M  \uu^j \mathbf{\Omega}^{-1} (\uu^j)^\top = \operatorname{tr}(\UU\mathbf{\Omega}^{-1} \UU^\top)$.
Given an instance $\langle x_i, c_{ki}\rangle$, the gradient with respect to $\uu_k$ is
\begin{equation}\label{update_u}
\begin{split}
     \frac{\partial \mathcal O }{\partial \uu_k} 
     = & -\Big(c_{ki} - \sigma(\boldsymbol{\beta}_k^\top \xx_i) \Big) \frac{ \partial \boldsymbol{\beta}_k^\top \xx_i}{\partial \uu_k} 
     +  \left[\UU \mathbf{\Omega}^{-1}\right]_k
\end{split}
\end{equation}
Note that $\beta_{jk}= \tau \lambda_{jk} w_{jk}$ with $\lambda_{jk} = \mathrm{tan}\left(\frac{\pi \Phi(u_{jk})}{2}\right), \Phi(u_{jk}) = \frac {1}{2}\left[1+\operatorname {erf} \left({\frac {u_{jk}}{ \sqrt {2\Omega_{kk} }}}\right)\right]$. Then $\frac{\partial \boldsymbol{\beta}_k^\top \xx_i}{\partial \uu_k} = \tau \frac{\partial f(\uu_k)}{\partial \uu_k} \circ \xx_i$, where $\left.\frac{\partial f(\uu_k)}{\partial \uu_k}\right\vert_{jk} = \frac{\pi}{2}\mathrm{sec}^2\left(\frac{\pi \Phi(u_{jk})}{2}\right) \frac{1}{\sqrt {2\pi \Omega_{kk}^2}}\exp\left(-\frac{u_{jk}^2}{2\Omega_{kk}^2}\right)$.

\vspace{2pt}
{\noindent \bf Update $\tau$ given others}:
With other parameters fixed, the objective function w.r.t  $\tau$ becomes 
\begin{equation*}\label{Equ:obj_fun_tau}
\begin{split}
\underset{\tau}{\operatorname{arg\,min}} & \sum_{k=1}^K\sum_{i=1}^{\mathcal N_k} \bigg\{ -c_{ki} \log \sigma(\boldsymbol{\beta}_k^\top \xx_i) - (1-c_{ki}) \log(1- \sigma(\boldsymbol{\beta}_k^\top \xx_i)) \bigg\}\\
& + 2\log(1+\tau^2) 
\end{split}
\end{equation*}
The gradients with respect to $\tau$ are given by
\begin{equation}\label{update_lambda}
\begin{split}
     \frac{\partial \mathcal O }{\tau}
     = -\sum_{k=1}^K\sum_{i=1}^{\mathcal N_k} \Big(c_{ki} - \sigma(\boldsymbol{\beta}_k^\top \xx_i) \Big)  \boldsymbol{\lambda}_k^\top \xx_i
      + \frac{4\tau}{1+\tau^2}
\end{split}
\end{equation}
where $\boldsymbol{\lambda}_k$ is the $k$-column of $\mathbf{\Lambda}$.
This allows $\tau$ to be updated using gradient decent.

\vspace{2pt}
{\noindent \bf Update $\mathbf\Omega$  given others}:
With other parameters fixed, the objective function w.r.t  $\mathbf\Omega$ becomes 
\begin{equation*}\label{Equ:obj_fun_Omega1}
\begin{split}
\underset{\mathbf\Omega}{\operatorname{arg\,min}} &  \sum_{z=1}^Z \Bigg\{ \frac{1}{2}\mathrm{tr}\left[ \mathbf{\Omega}^{-1} \WW_z^\top \mathbf{\Sigma}_z^{-1} \WW_z  \right] \Bigg\} 
 +  \frac{\delta}{2} \operatorname{tr} ( \mathbf\Omega_0 \mathbf\Omega^{-1})  + \frac{\xi}{2} \log |\mathbf\Omega|,
\end{split}
\end{equation*}
where $\xi = 2M + K + \nu + 1$. 
The last term $\log |\mathbf{\Omega}|$ can be seen as a penalty on the complexity of $\mathbf{\Omega}$, and can be replaced with the constraint $\mathrm{tr}(\mathbf{\Omega}) = 1$~\cite{Zhang:MLT:UAI2010}. 
Then above Equation (\ref{Equ:obj_fun_Omega1}) can be reformulated as:
\begin{equation}\label{Equ:obj_fun_Omega_relax}
\begin{split}
\underset{\mathbf\Omega}{\operatorname{arg\,min}}~~ &  \sum_{z=1}^Z \Bigg\{ \frac{1}{2}\mathrm{tr}\left[ \mathbf{\Omega}^{-1} \WW_z^\top \mathbf{\Sigma}_z^{-1} \WW_z  \right] \Bigg\} 
+  \frac{\delta}{2} \operatorname{tr} ( \mathbf\Omega_0 \mathbf\Omega^{-1}) \\
\mathrm{s.t.}~~ & \mathbf{\Omega} \succeq 0, ~ \mathrm{tr}(\mathbf{\Omega}) = 1
\end{split}
\end{equation}
where $\mathbf{\Omega} \succeq 0$ means that the matrix $\mathbf{\Omega}$ is positive semidefinite.
Equation (\ref{Equ:obj_fun_Omega_relax}) has an analytical solution:
\begin{equation}\label{update_Omega}
\begin{split}
     \mathbf{\Omega}
     = \frac{ \bigg( \frac{1}{2} \sum_{z=1}^Z \WW_z^\top \mathbf{\Sigma}_z^{-1} \WW_z  + \frac{\delta}{2} \mathbf\Omega_0  \bigg)^{\frac{1}{2}} }
     {\mathrm{tr} \Bigg[ \bigg( \frac{1}{2} \sum_{z=1}^Z \WW_z^\top \mathbf{\Sigma}_z^{-1} \WW_z   + \frac{\delta}{2} \mathbf\Omega_0  \bigg)^{\frac{1}{2}} \Bigg] }.
\end{split}
\end{equation}

\vspace{5pt}
{\noindent \bf Update $\mathbf{\Sigma}_z$ given others}:
With other parameters fixed, the objective function w.r.t  $\mathbf\Sigma_z$ becomes 
\begin{equation*}\label{Equ:obj_fun_Sigma1}
\begin{split}
\underset{\mathbf{\Sigma}_z}{\operatorname{arg\,min}}~ &  \frac{1}{2}\mathrm{tr}\left[ \mathbf{\Omega}^{-1} \WW_z^\top \mathbf{\Sigma}_z^{-1} \WW_z  \right] + \frac{K}{2} \log |\mathbf{\Sigma}_z|.
\end{split}
\end{equation*}
Similar to the case of updating $\mathbf\Omega$, the last term $\log |\mathbf{\Sigma}_z|$ in Equation (\ref{Equ:obj_fun_Sigma1}) can be seen as a penalty on the complexity of $\mathbf{\Sigma}_z$, and can be replaced with a constraint $\mathrm{tr}(\mathbf{\Sigma}_z) = 1$. Then above Equation (\ref{Equ:obj_fun_Sigma1}) can be reformulated as:
\begin{equation}\label{Equ:obj_fun_Sigma_relax}
\begin{split}
\underset{\mathbf{\Sigma}_z}{\operatorname{arg\,min}} &  ~~\mathrm{tr}\left[ \mathbf{\Omega}^{-1} \WW_z^\top \mathbf{\Sigma}_z^{-1} \WW_z  \right] ~~
\mathrm{s.t.}  ~~ \mathbf{\Sigma}_z \succeq 0, ~ \mathrm{tr}(\mathbf{\Sigma}_z) = 1.
\end{split}
\end{equation}
The Equation (\ref{Equ:obj_fun_Sigma_relax}) has an analytical solution:
\begin{equation}\label{update_Sigma}
\begin{split}
     \mathbf{\Sigma}_z
     = \frac{ \Big( \WW_z \mathbf{\Omega}^{-1} \WW_z^\top   \Big)^{\frac{1}{2}} }
     {\mathrm{tr} \bigg[ \Big( \WW_z \mathbf{\Omega}^{-1} \WW_z^\top \Big)^{\frac{1}{2}} \bigg] }.
\end{split}
\end{equation}

\begin{table*}[th]
\addtolength{\tabcolsep}{0pt}
\caption{List of the T2DM complications and the number of subjects included in this study.}\label{label:T2DMcomp}
\vspace{-0.3cm}
\begin{center}
\rowcolors{2}{gray!25}{white}
 \small
    \begin{tabular}{  L{2.5cm}  L{9.0cm}  L{3.0cm} R{1.3cm}}
    \rowcolor{gray!50}
    \toprule
   Complication   & Description & Example ICD-9 codes  & \# Subjects \\ \hline 
   Retinopathy (RET)  & Eye disease caused by damage to the blood vessels in the tissue at the back of the eye (retina) &25050, 25052, 24950, 24951, 36201-36207 & 7552\\
   Neuropathy (NEU) & Nerve damage most often affecting the legs and feet &25060, 25062, 24960, 24961 & 11151  \\
   Nephropathy (NEP) & Kidney disease or damage &25040, 25042, 24940, 24941 & 3969 \\
   Vascular Disease  (VAS) & Vascular diseases including peripheral vascular disease, cardiovascular disease, and cerebrovascular diseases &25070, 25072, 24970, 24971 & 6735 \\
   Cellulitis (CEL) & Serious bacterial skin infection & 37313, 37531, 38010-38016 & 11148 \\
   Pyelonephritis (PYE) & Inflammation of the kidney, typically due to a bacterial infection & 5900 - 5909 & 609 \\
   Osteomyelitis (OST) & Inflammation or infection of the bone and bone marrow; common in patients with diabetic foot problems & 73000-73007, 73009-73017, 73019-73027, 73029 & 909 \\
   Renal (REN) &  Renal failure & 28521, 585, 5854-5856, 586, 5845-5849& 5172 \\
   Hyperosmolar state (HHS) & One of two serious metabolic derangements characterized by hyperglycemia, hyperosmolarity, and dehydration without significant ketoacidosis & 25020, 25022, 24920, 24921 & 1077 \\
   Ketoacidosis (KET) & A complex disordered metabolic state characterized by hyperglycemia, ketoacidosis, and ketonuria & 25010, 25012, 24910, 24911 & 1617 \\
   Sepsis (SEP) & Immune response triggered by an infection that causes injury to the body's own tissues and organs & 0380-0389 & 2559  \\
   Shock (SHK) & A critical condition brought on by a sudden drop in blood flow through the body & 78550, 78551, 78552, 78559 & 777  \\
    \bottomrule
    \end{tabular}
\end{center}
\vspace{-0.3cm}
\end{table*}

\section{Experiments}\label{sec:exp}

In this section we present empirical evaluations to carefully vet our model on patient level data extracted from a large real-world electronic medical claims database. %

\subsection{Experimental Setup and Data}

We conduct a retrospective cohort study using the MarketScan Commercial Claims and Encounter (CCAE) database from Truven Health\footnote{https://truvenhealth.com/}. 
The data on the patients are contributed by a selection of large private employers' health plans, as well as government and public organizations.  
We use a dataset of de-identified patients between the years 2011 and 2014. 
The patient cohort used in the study consisted of T2DM patients selected based on the following criteria:
\begin{enumerate}
\setlength\itemsep{0.1cm}
\item[I.] The frequency ratio between Type 2 diabetes visits to Type 1 diabetes visits is larger than $0.5$; AND  
\item[II-a.] The patient has two (2) or more Type 2 diabetes labeled events on different days; OR
\item[II-b.] The patient received insulin and/or antidiabetic medication.
\end{enumerate} 
We focus on the risk of developing complications in the two year time window immediately following the initial T2DM diagnosis. 
Guided by clinical experts and following the report from American Diabetes Association~\cite{american2003report}, we identified 12 common complications of T2DM. 
We selected patients with at least two years of observations before the initial T2DM diagnosis. 
Further, patients who were under 19 years or age or over 64 years or age at the initial T2DM diagnosis are removed. 
Table \ref{label:T2DMcomp} shows the complications selected in this study and the corresponding number of patients. \\ 

We use following prediction variables:
\begin{packeditemize}
\setlength\itemsep{0.1cm}
\item {\bf Patient demographics:} age and gender. 
\item {\bf Diagnoses:} historical medical conditions encoded as International Classification of Disease (ICD) codes. ICD codes are grouped according to their first three digits and ICD codes appearing in fewer than $200$ patients are filtered out. This results in $296$ unique ICD features. Patients with less than $10$ occurrences of ICD codes are removed.
\item {\bf Medications:} medications that were received before the initial T2DM diagnosis date. A total of $19$ therapeutic classes related to glucose control, cardiac related drugs, and antibiotics were selected.
\end{packeditemize}
This results in a total of $317$ features.

\begin{table*}[!t]
\addtolength{\tabcolsep}{0pt}
\caption{Performance comparisons between the proposed TREFLES model and the baseline approaches for the 12 complications. The AUC average and standard deviation (in parenthesis) over the 5-fold cross validation trials are reported.}\label{tabel:AUC}
\vspace{-0.4cm}
\begin{center}
\normalsize
    \begin{tabular}{  c  c c c c c c c c c c c c}
    \toprule
   Method   & RET & NEU & NEP & VAS  & CEL &  PYE & OST & REN & HHS & KET & SEP & SHK  \\ \midrule
   \multirow{2}{*}{STL} & 0.5397 & 0.5889 & 0.5905 & 0.6581 & 0.5983 & 0.6222 & 0.7574 & 0.7351 & 0.6186 & 0.6558 & 0.7611 & 0.7794\\
   & (0.0108) & (0.0092) & (0.0096) & (0.0096) & (0.0049) & (0.0263) & (0.0468) & (0.0110) & (0.0323) & (0.0240) & (0.0152) & (0.0410)\\ \midrule
   \multirow{2}{*}{MTFL} & 0.5487 & 0.6034 & 0.6340 & 0.7059 & 0.6047 & 0.5604 & 0.8094 & 0.7801 & 0.6794 & 0.7011 & 0.7962 & 0.8316 \\
   & (0.0073) & (0.0134) & (0.0086) & (0.0085) & (0.0077) & (0.0687) & (0.0565) & (0.0078) & (0.0311) & (0.0335) & (0.0099) & (0.0292)\\ \midrule
   \multirow{2}{*}{MTRL} & 0.5643 & 0.6100 & 0.6456 & 0.7069 & 0.6283 & 0.6909 & 0.8480 & 0.7933 & 0.6990 & 0.7347 & 0.8182 & 0.8679\\
   & (0.0087) & (0.0103) & (0.0099) & (0.0105) & (0.0046) & (0.0633) & (0.0534) & (0.0073) & (0.0187) & (0.0348) & (0.0178) & (0.0209)\\ \midrule
   \multirow{2}{*}{FETR} & 0.5815 & 0.6488 & 0.6336 & 0.7290 & 0.6589 & 0.6913 & 0.8610 & 0.8087 & 0.6878 & 0.7320 & 0.8298 & 0.8709\\
   & (0.0178) & (0.0063) & (0.0126) & (0.0137) & (0.0067) & (0.0474) & (0.0506) & (0.0163) & (0.0304) & (0.0416) & (0.0140) & (0.0262)\\ \midrule
    \multirow{2}{*}{\bf TREFLES} & {\bf 0.5985} & {\bf 0.6697} & {\bf 0.6655} & {\bf 0.7478} & {\bf 0.6793} & {\bf 0.7194} & {\bf 0.8828} & {\bf 0.8316} & {\bf 0.7229} & {\bf 0.7626} & {\bf 0.8425} & {\bf 0.8784} \\ 
    & (0.0150) & (0.0075) & (0.0130) & (0.0091) & (0.0074) & (0.0422) & (0.0571) & (0.0130) & (0.0242) & (0.0341) & (0.0165) & (0.0247)\\
    \bottomrule
    \end{tabular}
\end{center}
\vspace{-0.3cm}
\end{table*}

\subsection{Evaluation Protocol}

{\noindent \bf Baselines.} We compare the new {\bf TREFLES} method with following set of strong baselines:
\begin{packeditemize}
\item Single task learning ({\bf STL}): For each task, we use a logistic regression to model the risk of each complication independently.  
\item Multi-task feature learning ({\bf MTFL})~\cite{Argyriou:MLFT:nips2007,Argyriou:MLFT:MLJ2008}: MTFL assumes that task association is captured through a subset of features shared among tasks.  It learns a few features common across the tasks using group sparsity, \ie the $\ell_1/\ell_2$-norm regularization on $\WW$, which both couples the tasks and enforces sparsity.
\item Multi-task relationship learning ({\bf MTRL})~\cite{Zhang:MLT:UAI2010}: MTRL assumes that the task association is revealed in the structure of the coefficient matrix $\WW$, but it only considers the task correlations in $\WW$ neglecting the correlations between features.  
\item Feature and task relationship learning ({\bf FETR})~\cite{zhao2017FETR}: FETR learns the relationships both between tasks and between features simultaneously. It can be seen a special case of our model without feature grouping and correlated  shrinkage. 
\end{packeditemize}

{\noindent \bf Evaluation metrics.}
We evaluate the models using AUC (area under the receiver operating characteristic curve), which is a standard metric in predictive analytics. 

\vspace{5pt}
{\noindent \bf Training and testing.} %
We used 5-fold cross validation to report results for each model. All the  models are implemented with  gradient descent optimization and we apply the Adam~\cite{adam} method to automatically adapt the step size during parameter estimation. 

\subsection{Incorporating Domain Knowledge}
{\noindent \bf Grouping of features.}
We group ICD features according to the domain knowledge encoded in the ICD ontologies.
Specifically, we group ICD-9 codes together when they have a same parent node (3 digits) in the ICD-9 hierarchy. 

\vspace{3pt}
{\noindent \bf Prior risk association $\mathbf{\Omega}_0$.}
Note that our model can incorporate prior knowledge on complication associations through $\mathbf{\Omega}_0$. We construct prior associations using the human disease network~\cite{HumanDiseaseNetwork:PANS2007}, which provides the Phi-correlations between pairs of diseases. We aggregate the Phi-correlations between pairs of ICD codes under pairs of T2DM complications. This results in a $\mathbf{\Omega}_0$ that represents our prior knowledge about the correlations between the T2DM complications in our study.

\subsection{Results Comparison}
\begin{figure}[t]
\centering
\centering\includegraphics[width=0.4\textwidth]{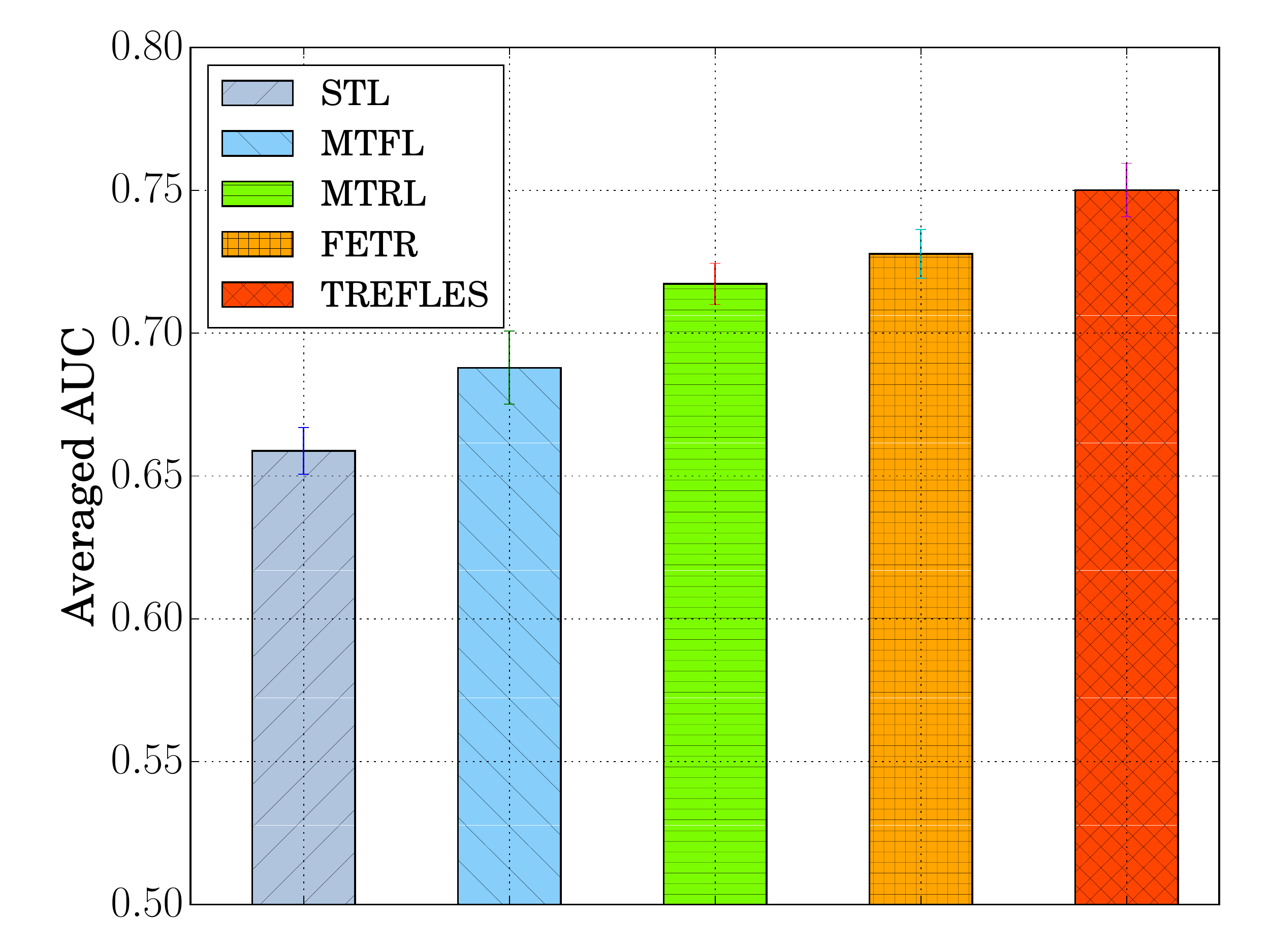}
\vspace{-0.4cm}
\caption{Performance comparisons between the proposed TREFLES model and the baseline approaches in terms of AUC (averaged over all 12 tasks).}\label{fig:means}
\vspace{-0.5cm}
\end{figure}

Table \ref{tabel:AUC} shows the AUCs between the proposed TREFLES model and the baseline approaches on all 12 complication risk prediction tasks. The average and standard deviation (in parenthesis) over the 5-fold cross validation trials are reported. 
Our approach consistently outperforms the baseline methods on all the 12 tasks.
Figure \ref{fig:means} plots the average AUCs and standard deviations across the 12 tasks for the different methods.

\vspace{3pt}
{\noindent \bf MTL versus STL:} We observe that all multi-task learning models (MTFL, MTRL, FETR and TREFLES) consistently and significantly outperform the single task learning method. In particular, our TREFLES model outperforms the single task learning method by $9.1\%$ in AUC on average. This confirms our assumption that directly modeling complications as independent of one another can lead to suboptimal models. Note that the different complications are manifestations of a common underlying condition--hyperglycemia, so their risks should be related. By simultaneously modeling multiple complications, MTL can capture and leverage the associations between complications using a shared representation. As a result, MTL models can significantly outperform STL models in risk prediction of T2DM complications.

\vspace{3pt}
{\noindent \bf TREFLES model versus baseline MTL models:} As shown in Figure \ref{fig:means}, our TREFLES model outperforms all baseline MTL models. TREFLES (AUC $0.7501 \pm 0.0091$) is better than the best baseline model FETR (AUC $0.7278 \pm 0.0094$) by $2.2\%$ in AUC. We also observe that the task relationship learning based method MTRL (AUC $0.7173 \pm 0.0072$) is more favorable than the feature relationship learning based method MTFL (AUC $0.6879 \pm 0.0128$). FETR outperforms MTRL because it simultaneously learns the relationships both between tasks and between features. TREFLES not only captures the relationships between tasks and between features, it also identifies the common contributing risk factors through the correlated coefficient shrinkage mechanism and incorporates domain knowledge through carefully constructed priors. As a result, TREFLES can significantly improve upon FETR.

\begin{figure}
\centering
\centering\includegraphics[width=0.5\textwidth]{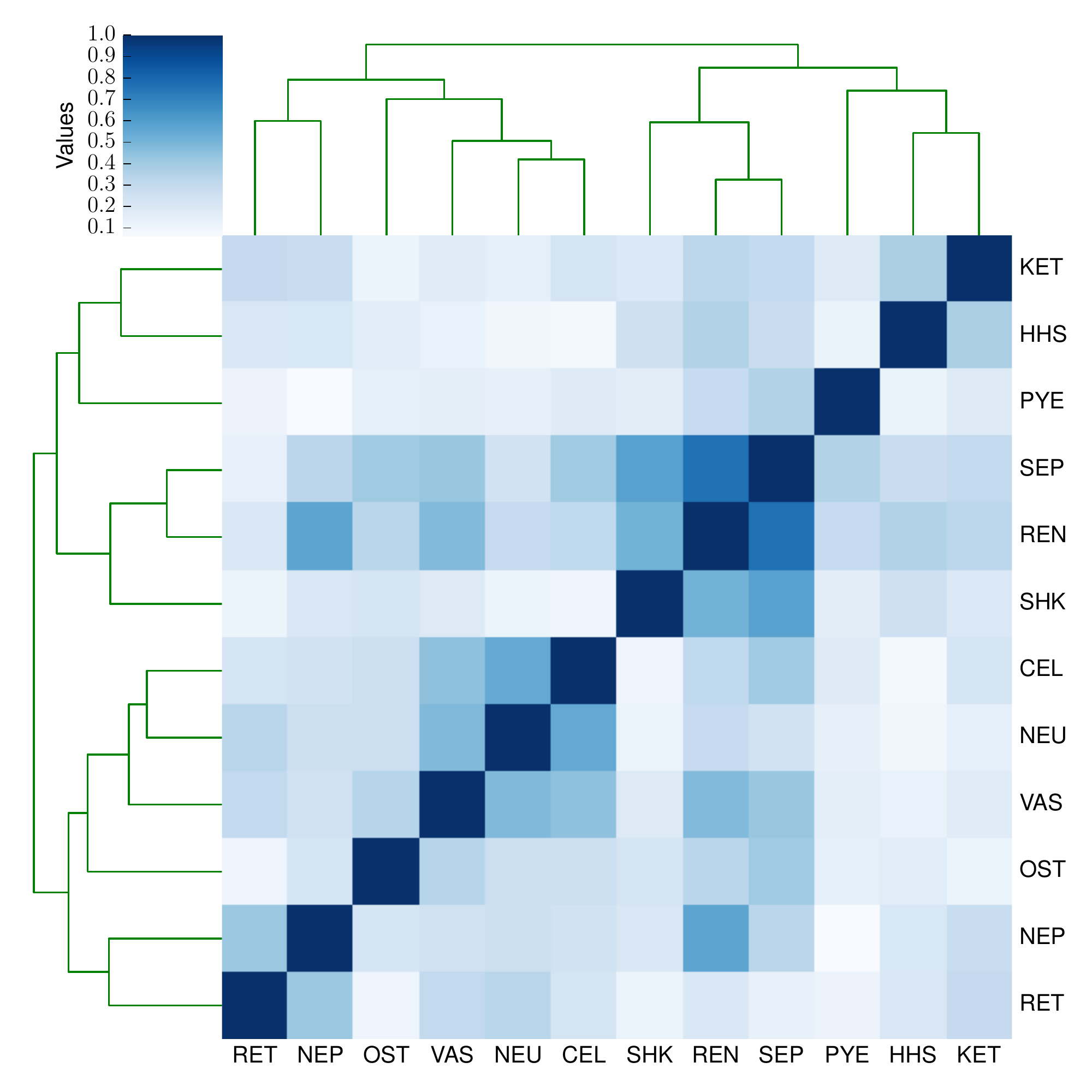}
\vspace{-0.7cm}
\caption{Heatmap and dendrogram of the hierarchical clustering of the  correlation matrix learned by TREFLES.}\label{fig:cor_heatmap}
\vspace{-0.6cm}
\end{figure}

\subsection{Learned Risk Associations}
In this section we discuss the estimated risk association matrix $\hat{\mathbf{\Omega}}$ from our TREFLES model. Matrix $\hat{\mathbf{\Omega}}$ represents the relatedness between complications learned from data. We first transfer the covariance matrix $\hat{\mathbf{\Omega}}$ to its correlation matrix $\hat{\mathbf{\RR}}$, whose elements have a ranges from $-1$ to $1$. We observe that all the elements in the correlation matrix $\hat{\mathbf{\RR}}$ learned by TREFLES have positive values. This is because all the complications are manifestations of a common underlying condition--hyperglycemia and they are positively correlated. 
Then we perform a hierarchical clustering on $\hat{\mathbf{\RR}}$. Figure \ref{fig:cor_heatmap} shows the heatmap and the dendrogram of the hierarchical clustering results. Darker colors indicate higher correlation. We can observe clusters between the risk associations of the 12 complications. In particular, CEL, NEU, VAS, OST, NEP and RET form one cluster while the remaining complications of KET, HHS, PYE, SEP, REN and SHK form a second cluster. 

The clusters are clinically meaningful. 
The first cluster of CEL, NEU, VAS, OST, NEP and RET represents the local complications caused by long standing or mismanaged diabetes, and the second cluster of KET, HHS, PYE, SEP, REN and SHK represents complications involving multiple sites or systemic complications.
Cluster 2 indicates more severe pathophysiologic manifestations of the disease than the cluster 1. 

\subsection{Identified Risk Factors}
Table \ref{label:RiskFactors} shows the top-5 risk factors/predictors (according to their coefficients) for each diabetic complication identified by our model. Most of the risk factors identified by our model are known to be clinically associated with the corresponding diabetic complications (indicated by *).  For example, the medical condition of ``Disorders of fluid, electrolyte, and acid-base balance'', which consistently appears in the top listing for all the diabetic complications, is indicative of many acid-based and electrolyte disorders that may be due to complications of T2DM and the medications diabetic patients receive.
Age is another major known risk factor for retinopathy, neuropathy, nephropathy and vascular disease including cardiovascular disease and the proposed method correctly identifies these associations. The underlying mechanism of age as a risk factor could be due to the fact that older adults tend to have long-standing diabetes, and consequently have associated microvascular and macrovascular complications.
Insulin treatment is identified as a risk factor for retinopathy, nephropathy, and cellulitis but not for the other complications. 

\begin{table*}[!t]
\addtolength{\tabcolsep}{3pt}
\caption{Top-5 risk factors (with the highest coefficients) for each complication as identified by the TREFLES model.}\label{label:RiskFactors}
\vspace{-0.3cm}
\begin{center}
\rowcolors{2}{gray!25}{white}
\footnotesize
    \begin{tabular}{  |L{5.3cm} | L{5.4cm} | L{5.3cm} |}
    \rowcolor{gray!50}
    \hline
      \multicolumn{1}{|c|}{Retinopathy (RET)} & \multicolumn{1}{c|}{Neuropathy (NEU)} & \multicolumn{1}{c|}{Nephropathy (NEP)} \\ \hline
      1.79 Antidiabetic Agents, Insulin*  & 4.07 {\footnotesize Hereditary and idiopathic peripheral neuropathy}
        & 1.98 {\scriptsize Disorders of fluid, electrolyte, and acid-base balance}*\\
      1.42 {\scriptsize Disorders of fluid, electrolyte, and acid-base balance}* & 4.03 Inflammatory and toxic neuropathy 
      &  1.29 Heart failure\\
      1.17 Other retinal disorders* & 2.42 Chronic ulcer of skin &  1.26 Antidiabetic Agents, Insulin\\
      1.12 Age* & 2.07 {\scriptsize Disorders of fluid, electrolyte, and acid-base balance}* & 1.26 {\footnotesize Nonspecific findings on examination of urine}\\
      0.89 {\footnotesize Nonspecific findings on examination of urine} & 1.70 Age* &  0.94 Age*\\ \hline
      \multicolumn{1}{|c|}{Vascular Disease (VAS)} & \multicolumn{1}{c|}{Cellulitis (CEL)} & \multicolumn{1}{c|}{Pyelonephritis (PYE)} \\ \hline
      8.32 Chronic ulcer of skin  & 3.89 Chronic ulcer of skin* & 1.65 {\scriptsize Disorders of fluid, electrolyte, and acid-base balance}\\
      3.10 {\scriptsize Disorders of fluid, electrolyte, and acid-base balance}* & 2.78 {\scriptsize Disorders of fluid, electrolyte, and acid-base balance} & 1.51 {\footnotesize Other disorders of urethra and urinary tract}* \\
      2.18 Hereditary and idiopathic peripheral neuropathy & 2.51 Bacterial infection in conditions classified elsewhere and of unspecified site* & 1.22 Bacterial infection in conditions classified elsewhere and of unspecified site* \\
      2.16 Age* & 2.20 Antidiabetic Agents, Insulin & 1.11 Calculus of kidney and ureter* \\
      1.88 Atherosclerosis* & 2.17 {\footnotesize Hereditary and idiopathic peripheral neuropathy}* & 0.91 Congenital anomalies of urinary system  \\ \hline
      \multicolumn{1}{|c|}{Osteomyelitis (OST)} & \multicolumn{1}{c|}{Renal (REN) } & \multicolumn{1}{c|}{Hyperosmolar state (HHS)} \\ \hline
      3.73 Chronic ulcer of skin* & 8.23 {\scriptsize Disorders of fluid, electrolyte, and acid-base balance}* & 4.40 {\scriptsize Disorders of fluid, electrolyte, and acid-base balance} \\
      1.84 Bacterial infection in conditions classified elsewhere and of unspecified site* & 3.04 Heart failure & 1.52 Heart failure*\\
      1.56 Open wound of foot except toes alone* & 2.71 Hypertensive chronic kidney disease* & 1.34 Disorders of mineral metabolism \\
      1.44 {\scriptsize Disorders of fluid, electrolyte, and acid-base balance} & 2.55 Chronic ulcer of skin & 1.25 Nondependent abuse of drugs*\\
      1.37 {\scriptsize Other and unspecified protein-calorie malnutrition} & 2.25 Other diseases of lung & 1.19 Hypertensive chronic kidney disease \\
      \hline
      \multicolumn{1}{|c|}{Ketoacidosis (KET) } & \multicolumn{1}{c|}{Sepsis (SEP) } & \multicolumn{1}{c|}{Shock (SHK) } \\ \hline
      5.68 {\scriptsize Disorders of fluid, electrolyte, and acid-base balance}* & 6.10 {\scriptsize Disorders of fluid, electrolyte, and acid-base balance}* & 6.10 {\scriptsize Disorders of fluid, electrolyte, and acid-base balance}*   \\
      1.23 Disorders of mineral metabolism & 3.46 Bacterial infection in conditions classified elsewhere and of unspecified site*
      & 2.42 Other diseases of lung\\
      1.22 {\footnotesize Nonspecific findings on examination of urine}*  & 3.39 Chronic ulcer of skin* & 2.06 Heart failure* \\
      1.10 Diseases of pancreas & 2.96 Other diseases of lung & 1.65 Pneumonia, organism unspecified \\
      1.03 Nondependent abuse of drugs* & 2.43 Chronic kidney disease (CKD)& 1.55 {\footnotesize Certain adverse effects not elsewhere classified}*\\
     \hline
     \multicolumn{3}{l}{ * indicates that the medical conditions have been mentioned in the clinical literature as the risk factors for the corresponding complications.}\\
    \end{tabular}
\end{center}
\vspace{-0.4cm}
\end{table*}

\section{Related Work}
From an applications perspective, our work falls into the category of studies that apply predictive analytics and use longitudinal patient records to improve the practice of healthcare management. Building predictive models from EHRs  and electronic medical claims data have attracted significant attention from both academia and industry, and have been applied to disease onset prediction~\cite{Kenney2016early,razavian2015population,himes2009prediction,cheng2016risk,choi2017using}, disease progression~\cite{wang:progression:kdd2014}, patient stratification~\cite{wang2015towards,chen2016patient}, hospital readmission prediction~\cite{he2014mining,bardhan2014predictive}, and mortality prediction~\cite{tabak2013using,Nori:KDD2015}. 
More recently, there have been some work focusing on diabetes. Razavian \etal~\cite{razavian2015population} show that claims data can be leveraged to predict T2DM onset. Oh \etal~\cite{oh2016type} applied EHRs to capture the trajectories of T2DM patients and found that different trajectories can lead to different risk patterns. Liu \etal~\cite{Liu:T2DM:aaai18} applied survival analysis to predict the onset of T2DM complications.  
Yadav \etal~\cite{yadav2015mining} presents a comprehensive survey on EHR data mining. 
Different from previous studies, this paper presents a comprehensive study to investigate the risk prediction and profiling of T2DM complications from patient medical records for diabetes care through a novel multi-task learning model.

Our work is also related to multi-task learning (MTL)~\cite{Caruana:MLT1997}, which aims to jointly learn multiple tasks using a shared representation so that knowledge obtained from one task can help other tasks. Recently, MTL models have been widely used in the healthcare domain~\cite{Zhou:KDD2011,Nori:KDD2015,Sun:KDD2015,wiens2016patient,Liu:T2DM:aaai18}. Feature relationship learning based approaches (known as MTFL)~\cite{Argyriou:MLFT:nips2007,Argyriou:MLFT:MLJ2008} and task relationship learning based approaches (known as MTRL)~\cite{Zhang:MLT:UAI2010} are the two most widely used MTL strategies~\cite{zhang:mtl_survey:2007}. MTFL assumes that task association is released through a subset of features shared among tasks. MTRL assumes that the task association is revealed in the structure of the coefficient matrix. Most similar to our approach is the feature and task relationship learning (FETR) method recently proposed by Zhao \etal~\cite{zhao2017FETR}. Similar to FETR, our proposed TREFLES model is a generalization of both MTRL and MTFL, and simultaneously learns the relationships both between tasks and between features. In healthcare analytics, associations between features are usually not ignorable. Different from FETR, TREFLES captures more fine-grained feature relationships by grouping features into groups according to domain knowledge. Furthermore, TREFLES is able to capture the correlated coefficient shrinkage among tasks through a novel correlated Horseshoe prior.  
As shown in our study, TREFLES is favorable for healthcare applications where we not only obtain better prediction performances, but also derive clinically meaningful insights about the relationships among the different complications and among the different risk factors.

\section{Conclusion}
In this paper, we provided a systematic study on risk profiling by simultaneously modeling multiple complications in chronic disease care using T2DM as a case study. 
We proposed a novel multi-task learning model, \emph {TREFLES}, that jointly captures relationships between risks, risk factors, and risk factor selection learned from the data with the ability to incorporate domain knowledge as priors.
TREFLES is favorable for healthcare applications because in additional to improved prediction performance, clinically meaningful insights about the relationships among different complications and risk factors can also be identified.
Extensive experiments on a T2DM patient dataset extracted from a large electronic medical claims database validated the improved prediction performance of TREFLES over current state of the art methods.
Also the risk associations learned as well as the risk factors identified by TREFLES lead to meaningful insights that were consistent with clinical findings.

There are a number of interesting future research directions. 
First, different complications could correspond to different severities of diabetes and we can use this knowledge to impose additional constraints on the risk correlations to potentially improve performance.
Second, the coefficient shrinkage strategy can be extended to incorporate domain knowledge about the risk factors to potentially improve interpretability.
Finally, we are also interested in applying our model to other chronic disease conditions with multiple complications or comorbidities which might benefit from the proposed modeling innovations proposed here.


\bibliographystyle{abbrv}

\vspace{-0.2cm}
\bibliography{refs} 
\end{document}